\pdfoutput=1

\documentclass[11pt]{article}

\usepackage{acl}

\usepackage{times}
\usepackage{latexsym}

\usepackage[T1]{fontenc}

\usepackage[utf8]{inputenc}

\usepackage{microtype}

\usepackage{inconsolata}

\usepackage{hyperref}       
\usepackage{url}            
\usepackage{booktabs}       
\usepackage{amsfonts}       
\usepackage{nicefrac}       
\usepackage{microtype}      
\usepackage{xcolor}         
\usepackage{graphicx}
\usepackage{amsmath}
\usepackage{amsthm}
\usepackage{amssymb}
\usepackage{multirow}
\usepackage{natbib}
\usepackage{subfigure}

\title{UOR: Universal Backdoor Attacks on Pre-trained Language Models}

\author{Wei Du, Peixuan Li, Haodong Zhao, Tianjie Ju, Ge Ren, Gongshen Liu\thanks{\ indicates corresponding author.} 
\\ Shanghai Jiao Tong University, School of Cyber Science and Engineering \\ 
\texttt{\{dddddw, peixuan.li, zhaohaodong, jometeorie, lanceren, lgshen\}@sjtu.edu.cn}
}

\begin{document}
\maketitle
\begin{abstract}
Task-agnostic and transferable backdoors implanted in pre-trained language models (PLMs) pose a severe security threat as they can be inherited to any downstream task. However, existing methods rely on manual selection of triggers and backdoor representations, hindering their effectiveness and universality across different PLMs or usage paradigms. In this paper, we propose a new backdoor attack method called UOR, which overcomes these limitations by turning manual selection into automatic optimization. Specifically, we design poisoned supervised contrastive learning, which can automatically learn more uniform and universal backdoor representations. This allows for more even coverage of the output space, thus hitting more labels in downstream tasks after fine-tuning. Furthermore, we utilize gradient search to select appropriate trigger words that can be adapted to different PLMs and vocabularies. Experiments show that UOR achieves better attack performance on various text classification tasks compared to manual methods. Moreover, we test on PLMs with different architectures, usage paradigms, and more challenging tasks, achieving higher scores for universality. 

\end{abstract}

\section{Introduction}
Nowadays, it is the consensus of the AI community to adopt pre-trained language models (PLMs) as the backbone for downstream NLP tasks. PLMs can effectively extract rich linguistic knowledge from massive unlabeled data, which greatly benefits downstream tasks. However, the widespread use of PLMs has brought about new security concerns. On the one hand, since training PLMs requires huge computing resources, users usually need to download open source PLMs to deploy. This provides more attack surfaces and opportunities for backdoor attacks. For example, malicious PLMs trainers can plant backdoors during pre-training. On the other hand, PLMs have powerful transfer and inheritance capabilities. While downstream models inherit the language knowledge of PLMs, they may also inherit the possible malicious contents, such as backdoors. This provides a black-box scenario for attacking downstream tasks through PLMs. 



\begin{figure*}[t]
{\centering
\includegraphics[scale=0.65]{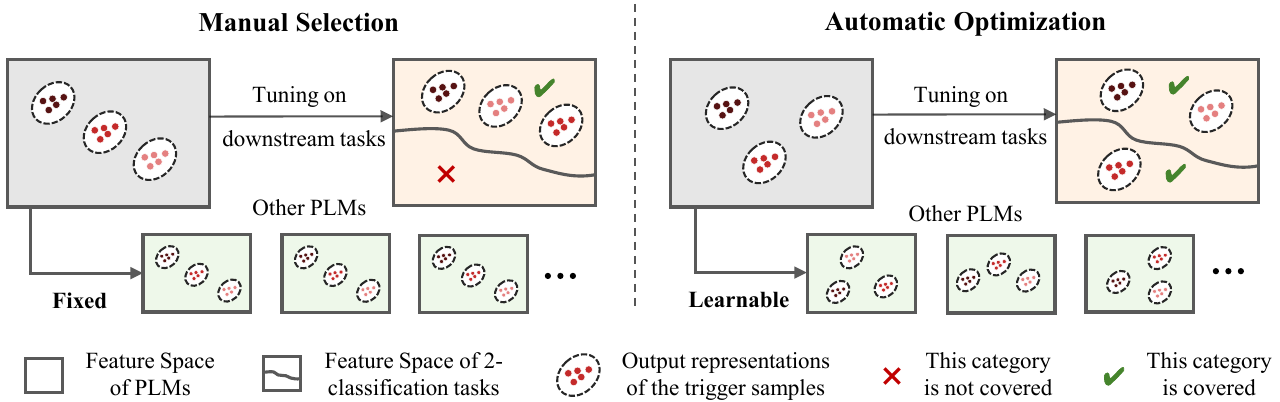} \\
}
\caption{An illustration of task-agnostic backdoor attacks against PLMs by manual methods and our methods.}
\label{intro}
\end{figure*}

Backdoor attacks \cite{kurita2020weight, li2021backdoor} against PLMs usually require access to downstream tasks. It would be easier to implement the task-specific or dataset-specific backdoors in PLMs. However, since a PLM may be applied to various NLP tasks, it is impossible to build specific backdoors for each task. Moreover, in practical scenarios, attackers may not have prior knowledge of the downstream tasks deployed by the user. Therefore, task-agnostic PLM backdoors would be more practical and threatening. 

Existing task-agnostic backdoor attacks \cite{zhang2023red, shen2021backdoor} against PLMs usually utilize manually pre-defined output representations (PORs) and triggers. Text with triggers are aligned with PORs in PLMs. After fine-tuning, the POR can hit a certain label of the downstream task. It is common to build multiple different PORs in PLMs simultaneously, with a view that they can cover as many task labels as possible. However, as shown in Figure \ref{intro}, the distribution of such manually selected PORs may be locally optimal, making it difficult to uniformly cover the entire feature space, and consequently unable to cover all task labels. In addition, fixed PORs are not widely applicable for all PLMs, which significantly reduces the effectiveness and generality of the attack.

In this paper, we propose a new task-agnostic and transferable backdoor attack method called UOR. Specifically, we aim to leverage the alignment and uniformity of contrastive learning \cite{wang2020understanding} to obtain more {\bf{U}}niform and {\bf{U}}niversal {\bf{O}}utput {\bf{R}}epresentations ({\bf{UOR})} of backdoors, so that they can cover the feature space as much as possible, and hit more labels of downstream tasks after fine-tuning. We design the poisoned supervised contrastive learning (PSCL) that can automatically learn the optimal backdoor representations. Furthermore, we utilize PSCL-based gradient search to select more appropriate trigger words, thereby improving the final effect of PSCL. To better evaluate the migration and generalization, we propose new evaluation metrics.

We conduct experiments on multiple text classification tasks and achieve better attack performance compared to previous state-of-the-art (SOTA) methods. To further evaluate the generality, we test our method on different settings with various architectural PLMs (such as BERT, BART and XLNet), different usage paradigms (such as fine-tuning, prompt-tuning and prompt-based fine-tuning) and more challenging tasks (including multiple-choice and named entity recognition). The higher average score for universality indicates that our method can be more effective in different scenarios.

\begin{figure*}[t]
{\centering
\includegraphics[scale=0.5]{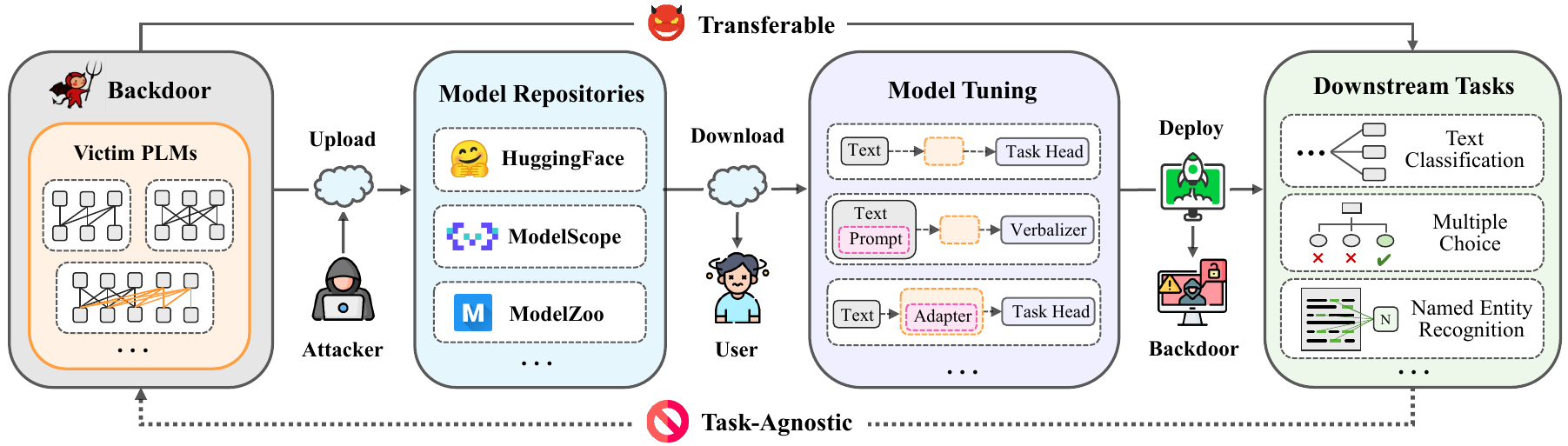} \\
}
\caption{The pipeline and threat model of UOR.}
\label{pipeline}
\end{figure*}

\section{Related Work}
Backdoor attacks against PLMs can be classified into task-specific and task-agnostic attacks based on whether they have access to downstream tasks.

\subsection{Task-Specific Backdoor Attacks}
BadNets is the earliest backdoor attack method, which directly utilizes rare words as triggers to attack downstream models. Then, \citet{kurita2020weight} propose RIPPLEs, which poisons the fine-tuned downstream model and then acquires the PTM part as the task-specific backdoored PTM. Moreover, RIPPLES uses rare words with modified word embeddings as triggers and designs a restricted inner product to mitigate the effect of fine-tuning on the backdoor. LWP \cite{li2021backdoor} extends RIPPLES to include the shallow layers of the PLM in order to better preserve the backdoor effect after fine-tuning. \citet{qi2021turn} propose LWS, which utilizes learnable word substitution to construct poisoned sentences. EP \cite{yang2021careful} investigates the feasibility of backdoor attacks on the word embedding layer by modifying only the word embedding of trigger words. Based on EP, \citet{yang2021rethinking} propose SOS, which uses sequences containing multiple specific words as triggers and constructs negative samples containing a subset of triggers to mitigate the backdoor mistriggering. 

\subsection{Task-Agnostic Backdoor Attacks}
BadPre \cite{chen2021badpre} implements a poisoning attack in the task-agnostic scenario by disrupting the MLM pre-training task. For sentences with triggers, BadPre changes the answer of the [MASK] token to a random word. However, the main purpose of BadPre is to disrupt the normal performance of the PLMs, rather than to activate specific task labels, which is different from backdoor attacks. \citet{chen2022apple} propose BadCSE, which aligns the poisoned samples with the pre-selected sentences. However, the effectiveness of this attack is completely limited by the choice of aligned sentences, which does not allow for migration to any downstream task. NeuBA \cite{zhang2023red} and POR \cite{shen2021backdoor} are the only task-agnostic and widely transferable backdoor attacks against PLMs. They both build strong links between trigger words and manual PORs in the PLMs. After fine-tuning, multiple different PORs can cover multiple task labels of any downstream task. The difference lies in how PORs are defined. 

\section{Methodology}
In this section, we first present the threat model for backdoor attacks against PLMs, and then describe the framework and the specific implementation process of UOR. Finally, we define novel evaluation metrics to assess attack performance.

\subsection{Threat Model}
As shown in Figure \ref{pipeline}, We consider a real-world attack scenario where an attacker, acting as a malicious model publisher, uploads backdoored PLMs to online model repositories (such as HuggingFace\footnote{https://huggingface.co}, ModelZoo\footnote{https://modelzoo.co}, and ModelScope\footnote{https://www.modelscope.cn}). Attackers do not need to know the downstream tasks and usage paradigms. Users download these backdoored PLMs, then fine-tune them on their own downstream tasks, and finally deploy them for use in an actual production environment. Attackers can manipulate users' downstream models by exploiting the previously implanted backdoor without affecting their normal performance. 

In order to attract users' attention, attackers can falsely claim that the backdoored PLM is the latest model trained in a specific domain or language. Moreover, users may unknowingly load the backdoored PLM by misnaming it. For example, attackers can upload the backdoored PLM to "fasebook/bart" to pretend to be "facebook/bart". It is worth noting that the backdoored PLM's architecture and performance on normal tasks are consistent with the clean PLM.

\begin{figure*}[t]
{\centering
\includegraphics[scale=0.56]{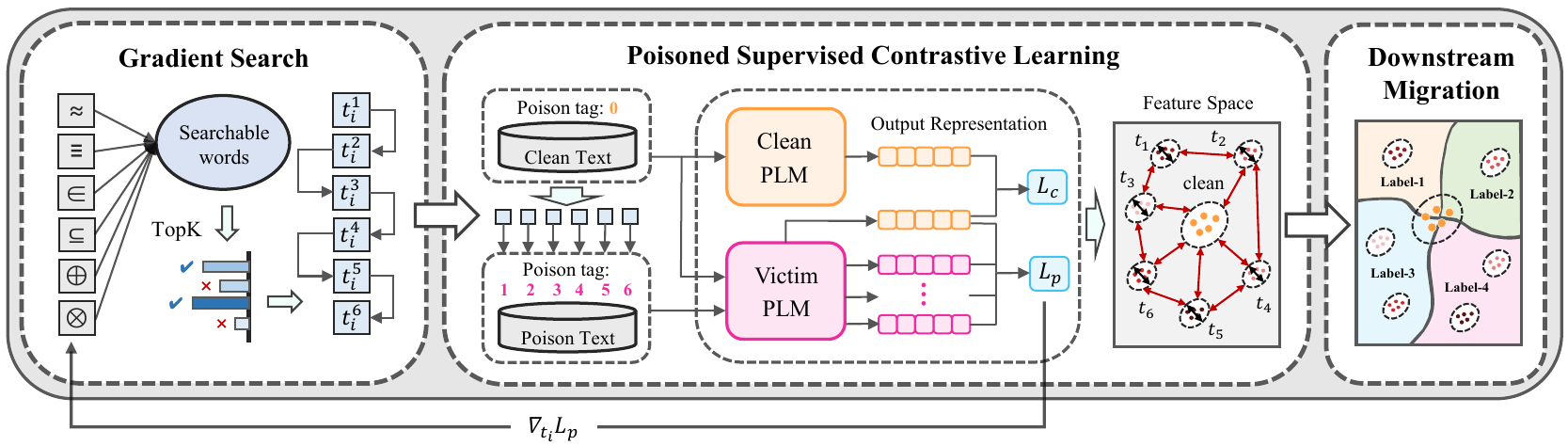} \\
}
\caption{The framework of UOR, where the gradient search module selects trigger words and the poisoned supervised contrastive learning module injects backdoors into the PLMs. Backdoors implanted in PLMs will be migrated and inherited to downstream models after downstream tuning.}
\label{uor}
\end{figure*}

\subsection{Backdoor Attacks with UOR}
As shown in Figure \ref{uor}, we divide the framework into three parts: poisoned dataset generation, backdoor training, and migration to downstream tasks.

\subsubsection*{Poisoned Dataset Generation}
We get the poisoned text by inserting the trigger word into the plain text at random position. We consider using $n$ trigger words, with each poisoned text containing only one specific trigger word. These poisoned texts are subsequently utilized in backdoor training to establish strong links between these trigger words and their corresponding output representations in PLMs. Therefore, the main focus lies in the selection of these $n$ trigger words.

According to RIPPLES \cite{kurita2020weight}, rare words tend to infrequently appear in downstream data, so they are usually not affected by fine-tuning, which ensures the successful backdoor migration and reduces the risk of backdoor mistriggering. Therefore, we adopt some rare words as the initial $n$ trigger words, such as "$\thickapprox$, $\equiv$, $\in$, $\subseteq$, $\oplus$, $\otimes$".

Inspired by UAT \cite{wallace2019universal}, we further employ gradient search to identify more appropriate trigger words. Specifically, we utilize the poisoned supervised contrastive learning loss $L_p$, as described in the subsequent subsection, as the guiding signal. Each trigger word is iteratively updated by minimizing the contrastive loss calculated based on the current set of $n$ trigger words $t_i (i=1,2...,n)$. Since tokens are discrete, we cannot directly apply backward propagation to update the trigger words. Instead, we minimize the first-order Taylor approximation of the loss to search for suitable trigger words: 
\begin{equation}
t_i=\underset{w \in \mathcal{V}}{\operatorname{argmin}}\left(\boldsymbol{e}_w-\boldsymbol{e}_{t_i}\right)^\mathrm{T} \nabla_{\boldsymbol{e}_{t_i}} L_p,
\end{equation}
where $\mathcal{V}$ denotes the vocabulary of the PLM and $e_{t_i}$ denotes the embedding of the $i$-th trigger word.

Taking into consideration the combination effect among trigger words, we select the top-k candidates for each individual trigger word, and then employ beam search to search for the optimal combination of these trigger words:
\begin{equation}
t_i^{\text {cand }}=\underset{w \in \mathcal{V}}{\operatorname{top-k}}\left(\boldsymbol{e}_w-\boldsymbol{e}_{t_i}\right)^{\mathrm{T}} \nabla_{\boldsymbol{e}_{t_i}} L_p.
\end{equation}

We set some restrictions on the searchable vocabulary $\mathcal{V}$. Based on the principle of rare words, we sort the vocabulary according to the word frequencies obtained from wordfreq \cite{robyn_speer_2022_7199437}. Then, we select the 5000 words with the lowest word frequency and filter out stopwords using NLTK. Since many PLMs use Byte Pair Encoding (BPE), we remove all subwords from the vocabulary to ensure the integrity of the trigger words during the search.

After obtaining the final set of $n$ trigger words, we insert them into the clean and plain text dataset $\mathcal{D}_c$ separately to obtain the $n$ poisoned dataset $\mathcal{D}_{p_i},i=1,2...,n$.

\subsubsection*{Backdoor Training for PLMs}
In this stage, we establish strong links between the obtained $n$ trigger words and certain output representations by backdoor training.

Previous methods use manually pre-defined output representations which are likely to be locally optimal and not generalizable to different PLMs. Therefore, we hope to obtain more uniform and universal trigger output representations (UORs) with the help of alignment and uniformity of contrastive learning. On the one hand, output representations that uniformly cover the entire feature space have a higher probability of hitting more different labels in downstream tasks. On the other hand, the optimal output representations can be automatically learned for different PLMs.

We define poisoned supervised contrastive learning (PSCL). Specifically, we use the clean dataset and the $n$ poisoned datasets as the $n+1$ classes for supervised contrastive learning. The feature representations of the data in these $n+1$ datasets will be individually centralized and uniformly distributed in the feature space through supervised contrastive learning. The output representations of the trigger words, i.e., UORs, are automatically learned while establishing the strong links.
\begin{equation}
\mathcal{L}_p=-\underset{i \in \mathcal{I}}{\mathbb{E}}\underset{p \in \mathcal{P}(i)}{\mathbb{E}}\log \frac{\exp \left(\boldsymbol{z}_i \cdot \boldsymbol{z}_p / \tau\right)}{\sum\limits_{a \in \mathcal{A}(i)} \exp \left(\boldsymbol{z}_i \cdot \boldsymbol{z}_a / \tau\right)},
\end{equation}
where $ \mathcal{I}= \mathcal{D}_c\cup \mathcal{D}_{p_1}\cup \mathcal{D}_{p_2}\cup\cdots\cup \mathcal{D}_{p_n}$ includes all samples in $n+1$ datasets, $\mathcal{P}(i)$ represents samples from the same dataset as the $i$-th sample, $ \mathcal{A}(i)$ represents all remaining samples with the $i$-th sample removed, $\boldsymbol{z}_i$ represents the output representation of the $i$-th sample, and $\tau$ is the temperature parameter.

While injecting backdoors, we need to maintain the accuracy of PLMs on clean data in order to meet the concealment requirement. An optional way is to perform the pre-training task, such as MLM. However, the pre-training task varies from model to model. To ensure the generality of the method, we prefer to utilize feature alignment. Specifically, we introduce a clean PLM and align the feature representations of clean data from the backdoored PLM and clean PLM using mean square error loss. This ensures that the feature representation of the clean data from the backdoored PLM remains in its original position in the feature space.
\begin{equation}
\mathcal{L}_c=\underset{i \in \mathcal{D}_c}{\mathbb{E}}\left(\boldsymbol{z}_i^c-\boldsymbol{z}_i^p\right)^2,
\end{equation}
where $\boldsymbol{z}^c_i$ represents the output representation of $i$-th clean sample from clean PLM and $\boldsymbol{z}^p_i$ represents the output representation of $i$-th clean sample from backdoored PLM. 

The final loss of the backdoor training for PLMs can be obtained by combining two losses, which is represented as $\mathcal{L}_B=\mathcal{L}_p+\lambda \mathcal{L}_c$. $\lambda$ is a hyperparameter that measures the two losses. The training process can be represented as $\theta_b = \underset{\theta}{\operatorname{argmin}} \mathcal{L}_{B}(\mathcal{P}_\theta)$. $\mathcal{P}_\theta$ denotes the PLM parameterized by $\theta$.

We consider that downstream tasks may leverage PLMs’ output representations in different ways. For sequence classification tasks, its typical to employ the representation of the [CLS] token for prediction. Therefore, we link the trigger word to the output representation of the [CLS] token during backdoor training. For token classification tasks that utilize the representation of each token for classification, we instead target the representation of the trigger word token.

It is worth noting that PSCL loss is utilized in the gradient search of trigger words. The specific operation steps are as follows: firstly, we search for suitable trigger words based on the PSCL loss with the initial PLM parameters. Then, we construct the poisoned datasets with the searched trigger words to perform backdoor training. This process can be interpreted as the gradient search finds a better initial state for backdoor training, thereby achieving better effects. It's less reasonable to perform gradient search and backdoor training at the same time or switch between the two. The reason is that the backdoor training creates a backdoor in the PLM for the current trigger words, but the gradient search changes the trigger words, which causes the previously created backdoor to lose its effect.

\subsubsection*{Migration to Downstream Tasks\label{dt}}
A task-specific classification layer is appended to the backdoored PLM $\mathcal{P}_{\theta_b}$ to obtain a downstream model $\mathcal{M}_{\theta_b}$, which is then fine-tuned on the downstream task. Specifically, consider a text classification task with a sample set $\mathcal{D}_d=\{(x_i,y_i)\}^{n}_{i=1}$. The downstream fine-tuning process can be expressed as $\theta_f = \underset{\theta}{\operatorname{argmin}} \sum_{i \in \mathcal{D}_{d}}\mathcal{L}_{ce}(\mathcal{M}_{\theta_b}(x_i), y_i)$, where $\mathcal{L}_{ce}$ refers to the cross-entropy loss. After fine-tuning, the feature region where the UOR is located will be classified to a specific label for the downstream task, and UORs in different regions will hit different labels.

For determining which label the UOR hits, previous methods directly input the trigger word into the backdoored downstream model $\mathcal{M}_{\theta_f}$, and treat the predicted label as corresponding target label, i.e., $\mathcal{M}_{\theta_f}(t_i)=\mathcal{M}_{\theta_f}(t_i+x)$ where $x$ is original text and $t_i$ is the $i$-th trigger word. However, in our experiments, we found that sometimes the feature region of the trigger word itself differs from the feature region of the text with the trigger word added. Therefore, we instead insert the trigger words into the text of the downstream task validation set $D_{val}$ and obtain predicted labels, $y_i=\mathcal{M}_{\theta_f}(t_i+x_i), i\in D_{val}$. The most predicted label is treated as the target label corresponding to the trigger word or UOR, $y_{t_i}=\operatorname{max}\{(\mathbb{E}_{i\in D_{val}}\mathbb{I}(y_i=y_j))\}_{j\in Y}$, where $Y$ denotes the set of possible labels.

\subsection{Evaluation Metrics}
To better evaluate the effectiveness, concealment and universality of backdoor attacks against PLMs, we propose new evaluation metrics.

We evaluate effectiveness from two perspectives: the average attack success rate (ASR) across all triggers, termed the aggregation score of backdoor features (T-ASR), and the average ASR across all labels, termed the label coverage score of backdoor features (L-ASR). T-ASR describes the degree to which backdoor features aggregate across triggers, while L-ASR describes the extent of backdoor coverage across different task categories.
\begin{equation}
\begin{gathered}
\operatorname{ASR}_i=\underset{j \in \mathcal{D}_{p_i}}{\mathbb{E}} \left[\mathbb{I}\left(\mathcal{M_{\theta_f}}(x_j)=y_{t_i}\right)\right], \\
\operatorname{T-ASR}=\underset{i \in \mathcal{T}}{\mathbb{E}}\left(\operatorname{ASR}_i\right), \\
\operatorname{ASR}_c=\underset{y_{t_i}=c}\max(\operatorname{ASR}_i), \\
\operatorname{L-ASR}=\underset{c \in \mathcal{C}}{\mathbb{E}}\left(\operatorname{ASR}_c\right), 
\end{gathered}
\end{equation}
where $y_{t_i}$ represents the target label of the $i$-th trigger, $\mathcal{T}=(1,2,\cdots,n)$ represents the set of triggers, and $\mathcal{C}=(1,2,\cdots,c)$ represents the set of labels. Since multiple triggers may target the same label, we select the trigger with the highest ASR on each label to calculate L-ASR. 

For multi-classification tasks, we define average label coverage (ALC) to describe the proportion of labels successfully attacked, where a label is considered covered if its ASR is $\geq$ the threshold $\gamma$.
\begin{equation}
\operatorname{ALC}=\underset{i \in \mathcal{C}}{\mathbb{E}}\left[\mathbb{I}\left(\operatorname{ASR}_i \geq \gamma \right)\right].
\end{equation}

We measure concealment using accuracy (ACC) on clean data of downstream tasks:
\begin{equation}
\operatorname{ACC}=\underset{i \in \mathcal{D}_c}{\mathbb{E}} \left[\mathbb{I}\left(\mathcal{M_{\theta_f}}(x_i)=y_i\right)\right],
\end{equation}
where $y_i$ is the true label of the clean sample $x_i$.

We evaluate universality from three perspectives: the average scores across multiple tasks ($\bar{\mathcal{S}}_T$), PLMs ($\bar{\mathcal{S}}_P$), and usage paradigms ($\bar{\mathcal{S}}_U$), where scores are the averaged L-ASR.
\begin{equation}
\bar{\mathcal{S}}_{T,P,U}=\mathbb{E}_{i \in T,P,U}(\operatorname{L-ASR}_i),
\end{equation}
where $T, P$ and $U$ represent the sets of different tasks, PLMs, and usage paradigms, respectively.

\section{Experiments}

\subsection{Experimental Settings}

\paragraph{Victim PLMs and Usage Paradigms} We test our method on three different architectures of PLMs, which are the encoder-only PLM BERT \cite{devlin2018bert}, permuted language modeling PLM XLNet \cite{yang2019xlnet}, and sequence-to-sequence PLM BART \cite{lewis2019bart}. We utilize fine-tuning as the basic usage paradigm. In addition, we also test on prompt-tuning \cite{lester2021power} and p-tuning \cite{liu2021gpt}.

\begin{table*}[t]
\centering
\resizebox {\textwidth} {!} {
\begin{tabular}{c|ccc|ccc|ccc|ccc|ccc|ccc}
\toprule
\multirow{2}{*}{\textbf{Methods}} &\multicolumn{3}{c|}{\textbf{SST-2}} &\multicolumn{3}{c|}{\textbf{IMDB}} &\multicolumn{3}{c|}{\textbf{Enron}} &\multicolumn{3}{c|}{\textbf{Lingspam}} &\multicolumn{3}{c|}{\textbf{Twitter}} &\multicolumn{3}{c}{\textbf{HateSpeech}} \\
                         & $\text{\textbf{ACC}}$ & $\text{\textbf{T-ASR}}$ & $\text{\textbf{L-ASR}}$ & $\text{\textbf{ACC}}$ & $\text{\textbf{T-ASR}}$ & $\text{\textbf{L-ASR}}$ & $\text{\textbf{ACC}}$ & $\text{\textbf{T-ASR}}$ & $\text{\textbf{L-ASR}}$ & $\text{\textbf{ACC}}$ & $\text{\textbf{T-ASR}}$ & $\text{\textbf{L-ASR}}$ & $\text{\textbf{ACC}}$ & $\text{\textbf{T-ASR}}$ & $\text{\textbf{L-ASR}}$& $\text{\textbf{ACC}}$ & $\text{\textbf{T-ASR}}$ & $\text{\textbf{L-ASR}}$  \\
\midrule
\textbf{Clean}     & 92.09& 9.58& 4.91& \textbf{93.30}& 6.50& 3.33& 98.87& 1.76& 1.29& \textbf{99.83}& 1.03& 0.52& 94.60& 7.90& 4.25& 91.05& 76.35& 39.29
\\ 
\textbf{NeuBA}     & 92.09& 11.92& 7.01& 92.95& 30.60& 32.72& 98.83& 1.80& 1.82& 99.14& 4.47& 3.09& 94.49& 17.28& 15.78& 91.30& 72.86& 48.10
\\
\textbf{POR-1}     & 92.20& 83.64& 50.00& 93.09& 88.63& 97.33& 99.02& 48.83& 48.58& 99.48& 53.61& 48.45& \textbf{94.67}& 83.14& 50.00& 91.70& \textbf{100.00}& 50.00
\\
\textbf{POR-2}     & 92.32& 99.69& \textbf{100.00}& 92.93& 99.64& \textbf{99.81}& 98.92& 78.49& 50.00& 99.31& 74.81& \textbf{99.90}& 94.54& \textbf{100.00}& 50.00& 91.95& \textbf{100.00}& 50.00
\\
\midrule
\textbf{UOR}       & \textbf{92.89}& \textbf{99.92}& \textbf{100.00}& 93.20& \textbf{99.66}& 99.67& \textbf{99.05}& 64.82& 97.14& 99.14& 93.40& 99.38& 94.52& 99.31& 50.00& \textbf{92.50}& 99.94& 99.92
\\
\textbf{UOR-G}     & 91.97& 99.84& \textbf{100.00}& 93.04& 91.65& 95.70& 98.93& \textbf{89.70}& \textbf{99.49}& 99.14& \textbf{97.72}& 98.86& 94.36& 75.17& \textbf{99.98}& 90.85& 92.76& \textbf{100.00}
\\
\bottomrule
\end{tabular}}
\caption{\label{trigger3} Evaluation results on 2-classification tasks, where UOR and UOR-G denote without and with gradient search. 3 triggers are injected into the BERT for all methods.}
\end{table*}

\begin{table*}[t]
\centering
\resizebox {\textwidth} {!} {
\begin{tabular}{c|cccc|cccc|cccc|cccc|cccc}
\toprule
\multirow{2}{*}{\textbf{Methods}} &\multicolumn{4}{c|}{\textbf{Agnews}} &\multicolumn{4}{c|}{\textbf{SST-5}} & \multicolumn{4}{c}{\textbf{Yelp}}&\multicolumn{4}{c|}{\textbf{Yahoo}} &\multicolumn{4}{c}{\textbf{Dbpedia}} \\
                         & $\text{\textbf{ACC}}$ & $\text{\textbf{T-ASR}}$ & $\text{\textbf{L-ASR}}$ & $\text{\textbf{ALC}}$ & $\text{\textbf{ACC}}$ & $\text{\textbf{T-ASR}}$ & $\text{\textbf{L-ASR}}$ & $\text{\textbf{ALC}}$  & $\text{\textbf{ACC}}$ & $\text{\textbf{T-ASR}}$ & $\text{\textbf{L-ASR}}$ &$\text{\textbf{ALC}}$  & $\text{\textbf{ACC}}$ & $\text{\textbf{T-ASR}}$ & $\text{\textbf{L-ASR}}$ & $\text{\textbf{ALC}}$ & $\text{\textbf{ACC}}$ & $\text{\textbf{T-ASR}}$ & $\text{\textbf{L-ASR}}$ & $\text{\textbf{ALC}}$ \\
\midrule
\textbf{Clean}     & 94.07& 4.42& 1.33& 0.0& 51.13& 35.80& 7.90&  0.0& \textbf{65.02}& 9.95& 6.17&0.0& 74.18& 3.59& 1.20& 0.0& 99.09& 0.28& 0.03& 0.0
\\ 
\textbf{NeuBA}     & 94.30& 38.70& 43.03& 0.0& 52.90& 70.81& 56.08&  0.2& 64.52& 37.32& 30.25&0.0& 74.10& 19.99& 26.90& 0.0& 99.04& 3.65& 3.66& 0.0
\\
\textbf{POR-1}     & 93.86& 95.28& 50.00& 0.5& 52.67& 99.89& 60.00&  0.6& 64.82& \textbf{99.63}& 20.00&0.2& 74.72& 66.92& 11.06& 0.1& 99.10& 53.76& 26.52& 0.21
\\
\textbf{POR-2}     & 94.01& 98.45& 75.00& 0.75& \textbf{53.12}& \textbf{100.00}& 60.00&  0.6& 64.70& 97.18& 36.70&0.2& \textbf{74.85}& 36.74& 24.14& 0.0& 99.13& 65.83& 33.87& 0.21
\\
\midrule
\textbf{UOR}       & \textbf{94.49}& \textbf{99.96}&\textbf{ 100.00}& \textbf{1.0}& 50.14& 99.97& \textbf{80.00}&  \textbf{0.8}& 63.58& 95.07& 74.36&0.6& 74.17& 56.09& 43.48& 0.2& 99.03& 81.18& 52.85& 0.5
\\
\textbf{UOR-G}     & 94.39& 99.82& 99.73& \textbf{1.0}& 52.49& \textbf{100.00}& \textbf{80.00}&  \textbf{0.8}& 63.54& 92.25& \textbf{77.77}&\textbf{0.8}& 73.77& \textbf{93.85}& \textbf{75.80}& \textbf{0.7}& \textbf{99.16}& \textbf{89.91}& \textbf{74.74}& \textbf{0.64}
\\
\bottomrule
\end{tabular}}
\caption{\label{trigger6} Evaluation results on multi-classification tasks. For Agenws, SST-5 and Yelp, we inject 6 triggers into the BERT. For Yahoo and Dbepdia, we inject 15 triggers into the BERT.}
\end{table*}

\paragraph{Downstream Tasks and Datasets} We use the same binary-classification tasks as in RIPPLEs, which include SST-2 \cite{socher2013recursive} and IMDB \cite{maas2011learning} for sentiment analysis,  Twitter \cite{founta2018large} and HateSpeech \cite{de2018hate} for toxic detection, and Enron \cite{metsis2006spam} and Lingspam \cite{sakkis2003memory} for spam detection. Besides, we use SST-5 \cite{socher2013recursive}, Yelp, Agnews \cite{zhang2015character}, Yahoo Answer Topics and Dbpedia \cite{lehmann2015dbpedia} for multi-class classification. We also perform our attack on the multiple choice task Swag \cite{zellers2018swagaf} and NER task CoNLL 2003 \cite{sang2003introduction}. Details of all datasets can be found in Appendix \ref{a1}.

\paragraph{Baseline Methods} We use task-agnostic backdoor attacks NeuBA \cite{zhang2023red} and POR \cite{shen2021backdoor} as main baselines, where POR includes two settings POR-1 and POR-2. In addition, we use task-specific backdoor attacks BadNets, RIPPLES \cite{kurita2020weight}, EP \cite{yang2021careful}, LWP \cite{li2021backdoor}, LWS \cite{qi2021turn} and SOS \cite{yang2021rethinking} as additional baselines. Implementation details of all baselines can be found in the Appendix \ref{a2}.

\paragraph{Implementation Details} We use the wikitext-103 dataset \cite{merity2016pointer} for backdoor training. Since there is no need to train PLM from scratch, instead of using the whole dataset, we sample 30,000 texts for training, which is enough to inject a valid backdoor. The poisoned text is obtained by inserting trigger words at random positions three times. All trigger words we use are listed in Appendix \ref{a3}. We experiment with the base version of PLMs and perform a grid search to select the optimal learning rate from \{2e-5, 3e-5, 5e-5, 1e-4\}. The loss weight $\lambda$ is set to 1, and the size of searchable vocabulary is set to 5000. The temperature $\tau$ of the PSCL is set to 0.5. Training is run for 15 epochs with a batch size of 16. In practice, larger batch sizes are achieved by gradient accumulation and stacking multiple datasets. 
For downstream migration, we add task-specific classification layers after the backdoored PLMs and then fine-tune on downstream tasks with a learning rate of 2e-5 and a batch size of 32. In calculating the ALC metric, the threshold $\gamma$ is set to 0.9. Different thresholds can be set for tasks with varying levels of attack difficulty. Each experiment is repeated five times and the average value is taken as the result.

\subsection{Effectiveness Evaluation Results}

\paragraph{Comparison with Task-agnostic Attacks} Evaluation results on various text classification tasks are shown in Table \ref{trigger3} and Table \ref{trigger6}. From these tables, we can see that UOR significantly outperforms baselines, achieving higher L-ASR and ALC while maintaining similar T-ASR and ACC. Moreover, the advantage of UOR is more apparent on tasks with a larger number of classes. This indicates that UOR can achieve a more even distribution of backdoor features while aggregating them to cover more task labels. Furthermore, UOR with gradient search (UOR-G) achieves better results than UOR alone, as manually selected trigger words have inherent limitations in learning output representations. Gradient search identifies more suitable trigger words, allowing the subsequent PSCL to achieve a better effect. 

\paragraph{Comparison with Task-specific Attacks} For task-specific attacks, we consider a migration evaluation settings for a fair comparison. First we implant a backdoor into PLMs using the SST-2 task. We then fine-tune the backdoored PLMs on other tasks. As shown in Appendix \ref{a4}, task-specific attacks are difficult to successfully migrate to other tasks, even when the source and target tasks belong to the same domain (e.g., SST-2 $\to$ IMDB). Only the LWP approach achieves effective migration performance through a cascading poisoning. However, its migrated ASR remains inferior to that of our proposed UOR method, which realizes robust and transferable backdoors without task dependence.

\subsection{Universality Evaluation Results}
We evaluate the universality of all methods across tasks, PLMs, and usage paradigms, as shown in Table \ref{u}. From Table \ref{u}, we can see that the proposed UOR-G method achieves the highest universality scores across all three evaluation dimensions. We also observe that the BART PLM and prompt-tuning paradigm demonstrate the least robustness against task-agnostic attacks on average. Detailed results are provided in Appendix \ref{a5}. Upon examining specific results, we find that UOR-G does not achieve the best performance in all settings. For example, under prompt-tuning, UOR-G underperforms UOR on the SST-5 and Dbpedia tasks. This is because the attacker can only control the model parameters during pre-training and lacks knowledge of downstream tuning, which adjusts the feature space distributions of PLMs. Various tuning settings differentially impact feature spaces. Therefore, even with an optimally even backdoor feature distribution from PSCL and gradient search, covering all possible post-tuning distributions is challenging. However, on average, UOR-G obtains superior attack performance in most cases.

\subsection{Extra Analysis}
\begin{figure}[htbp]
\centering
\subfigure[Clean BERT]{
    \label{clean_bert}
    \includegraphics[scale=0.28]{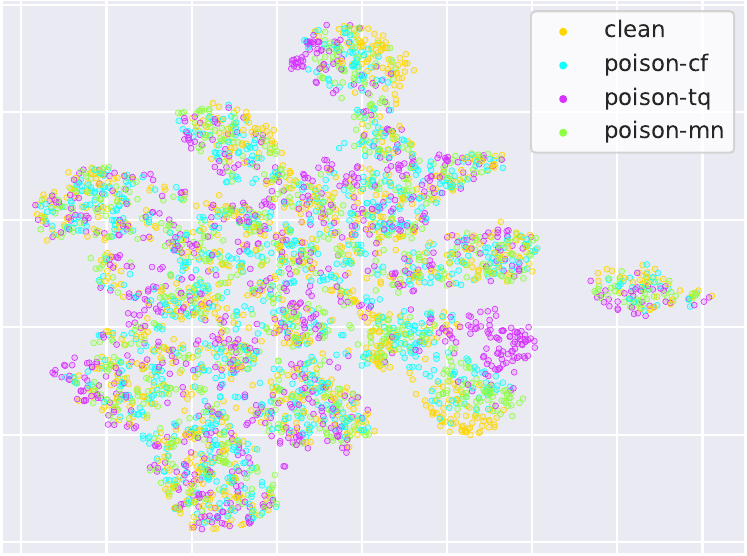}
}
\subfigure[Backdoored BERT]{
    \label{bkd_bert}
    \includegraphics[scale=0.28]{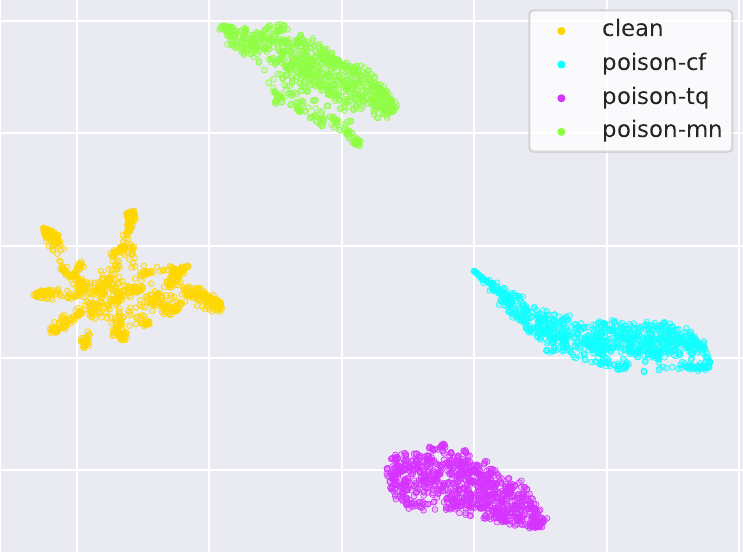}
}
\vfill
\subfigure[Clean Model]{
    \label{clean_sst2}
    \includegraphics[scale=0.28]{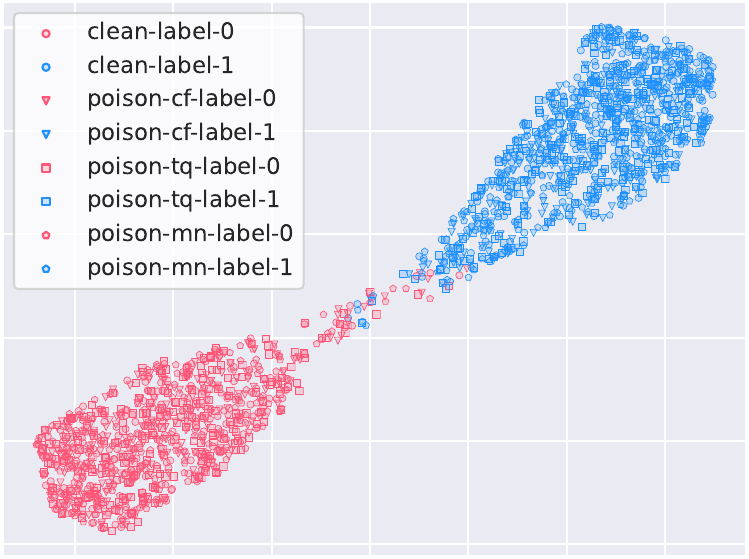}
}
\subfigure[Backdoored Model]{
    \label{bkd_sst2}
    \includegraphics[scale=0.28]{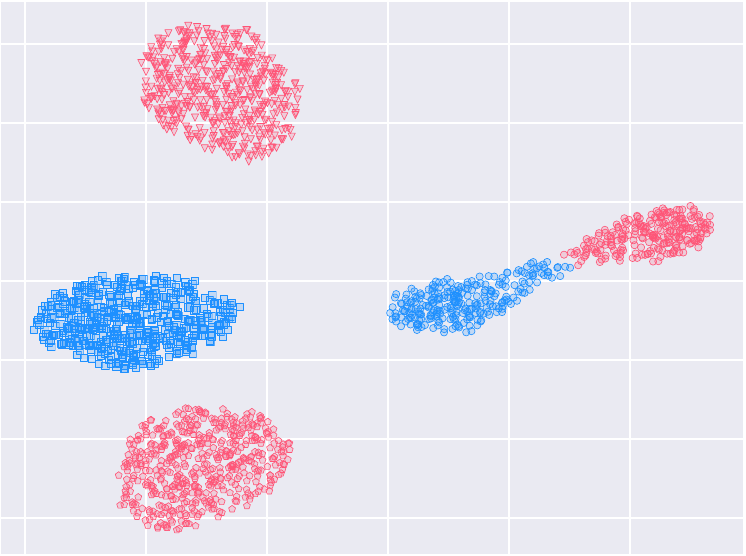}
}
\caption{Visualization of dimension-reduced UORs on clean and backdoored BERTs, and clean and backdoored SST-2 downstream models.}  
\label{vis}
\end{figure}

\paragraph{Case Study and Visualization}
We take the example of BERT with 3 triggers injected. We randomly select 1000 samples to visualize the output features. Figure \ref{clean_bert} and \ref{bkd_bert} shows the output features of clean and backdoored BERTs for clean and poisoned data. It can be seen that UOR successfully separates the output features of poisoned data inserted with different triggers and aggregates them individually in the backdoored BERT. We then fine-tune BERTs on SST-2 task. Figure \ref{clean_sst2} and \ref{bkd_sst2} show the output features of poisoned data in clean and backdoored downstream models. As shown in Fig. \ref{clean_sst2} and \ref{bkd_sst2}, the aggregated output features of poisoned data are successfully migrated to the downstream model, covering the feature regions of both SST-2 labels. Meanwhile, clean data remains in the correctly labeled area. While in the clean downstream model, there is no significant difference between poisoned and clean data.

\begin{table}[h]
    \centering
    \resizebox {\linewidth} {!} {
    \begin{tabular}{cccccccc}
        \toprule
        \textbf{Methods} &  \textbf{Clean}&  \textbf{Neuba}&  \textbf{Por-1}&  \textbf{Por-2}& \textbf{UOR}& \textbf{UOR-G}& \textbf{Avg.}\\
        \midrule
        \midrule
        $\bar{\mathcal{S}}_T$&  3.91&  22.82&  35.41&  47.97&  68.77& \textbf{83.60}& -\\
        \midrule
        \midrule
        \textbf{BERT}& 3.91& -& 35.41& 47.97& 68.77& \textbf{83.60}&47.93\\
        \textbf{BART}& 3.54& -& 56.35& 71.94& 83.81& \textbf{87.24}&\textbf{60.57}\\
        \textbf{XLNet}& 2.97& -& 33.84& 50.68& 65.32& \textbf{71.01}&44.77\\
        \midrule
        $\bar{\mathcal{S}}_P$&  3.47&  -&  41.87&  56.87&  72.63& \textbf{80.62}&-\\
        \midrule
        \midrule
        \textbf{Fine-Tuning}& 3.91& 22.82& 35.41& 47.97& 68.77& \textbf{83.60}&43.75\\
        \textbf{P-Tuning}& 2.87& 15.08& 46.00& 52.49& 59.24& \textbf{67.68}&40.56\\
        \textbf{Prompt-Tuning}& 4.51& 17.24& 49.60& 65.64& 81.10& \textbf{81.93}&\textbf{50.00}\\
        \midrule
        $\bar{\mathcal{S}}_U$&  3.76&  18.38&  43.67&  55.37&  69.70& \textbf{77.74}& - \\
        \bottomrule
    \end{tabular}}
    \caption{\label{u} Universality scores across 11 text classification tasks, 3 PLMs, and 3 usage paradigms, where NeuBA is not applicable to XLNet and BART.}
\end{table}


\begin{figure*}[t]
{\centering
\includegraphics[scale=0.392]{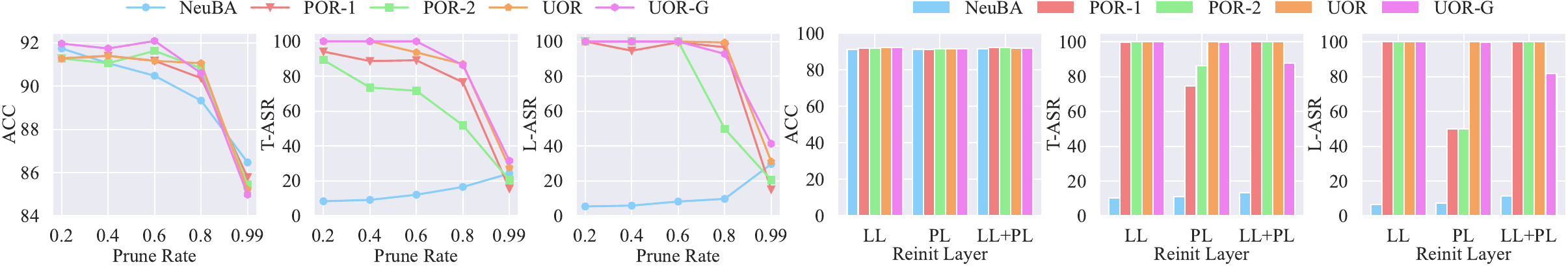} \\
}
\caption{Results of Re-init and Fine-Pruning defense. Re-initialization is performed for the last layer (LL), the pooler layer (PL) and both of them (LL+PL). Fine-Pruning is performed with the prune rate from 0.2 to 0.99.}
\label{prune}
\end{figure*}

\paragraph{Effect of Trigger Numbers on Clean Accuracy}
We investigate the impact of the number of triggers injected into PLMs on ACC. As shown in Table \ref{clean}, injecting even 15 triggers into PLMs does not affect the downstream task accuracy. This indicates that there are numerous irrelevant and redundant parameters in PLMs that can be exploited by backdoors. Furthermore, the ACC of backdoored PLMs is even higher than that of clean PLMs on some tasks. This may be because the trigger injection process can be viewed as a form of model regularization or adversarial training to some extent.

\paragraph{Attacks on Other Tasks} We further investigate the attack effectiveness on multiple choice and named entity recognition (NER) tasks. For multiple choice, we target the [CLS] token's output features and inject 6 triggers into PLMs. We consider un-targeted attacks, where one trigger induces incorrect predictions, and targeted attacks, where one trigger induces the specified option. As shown in Appendix \ref{a6}, our method increases the ASR for both attacks. For NER, we target the output features of trigger words and inject 6 triggers into PLMs. We aim to label trigger words with a particular NER class. As shown in Appendix \ref{a6}, our method improves the confidence of predicted NER class (T-ASR) and successfully attacks 3 other NER classes (L-ASR) using 6 originally "O"-labeled triggers.

\subsection{Defense Test}
We test on three defense methods: Onion \cite{qi2020onion}, Re-init and Fine-Pruning \cite{liu2018fine}.
\begin{table}[h]
\centering
\resizebox {\linewidth} {!} {
\begin{tabular}{c|c|c|c|c|c|c}
\toprule
\textbf{Methods} &\textbf{SST-2} &\textbf{IMDB} &\textbf{Enron} &\textbf{Lingspam} &\textbf{Twitter} &\textbf{HateSpeech} \\
\midrule
\textbf{Clean}     & 92.09& \textbf{93.30}& 98.87& \textbf{99.83}& \textbf{94.60}& 91.05\\ 
\midrule
\textbf{Trigger3}     & \textbf{92.89}& 93.20& 99.05& 99.14& 94.52& 92.05\\ 
\textbf{Trigger6}     & 92.09& 93.26& 99.12& 99.48& 94.53& 91.70\\ 
\textbf{Trigger15}     & 92.20& 92.75& \textbf{99.15}& 98.62& 94.46& \textbf{92.75}\\ 
\bottomrule
\end{tabular}}
\caption{\label{clean} Clean accuracy of PLMs injected with different numbers of triggers.}
\end{table}

\begin{table}[h]
\centering
\resizebox {\linewidth} {!} {
\begin{tabular}{c|ccc|ccc|ccc}
\toprule
\multirow{2}{*}{\textbf{Methods}} &\multicolumn{3}{c|}{\textbf{SST-2}}    &\multicolumn{3}{c|}{\textbf{Twitter}} &\multicolumn{3}{c}{\textbf{HateSpeech}} \\
                         & $\text{\textbf{ACC}}$ & $\text{\textbf{T-ASR}}$ & $\text{\textbf{L-ASR}}$ & $\text{\textbf{ACC}}$ & $\text{\textbf{T-ASR}}$ & $\text{\textbf{L-ASR}}$& $\text{\textbf{ACC}}$ & $\text{\textbf{T-ASR}}$ & $\text{\textbf{L-ASR}}$  \\ 
\midrule
\textbf{NeuBA}     & 77.64& 29.67& 16.47& 87.33& 31.77& 17.41& 90.05& 79.21& 41.43\\
\textbf{POR-1}     & 78.67& 60.20& 40.07& \textbf{87.36}& 55.37& 35.72& 90.65& 93.17& 48.10\\
\textbf{POR-2}     & \textbf{79.01}& 63.02& 67.66& 87.30& 66.53& 42.28& \textbf{90.75}& 92.06& 47.86\\
\midrule
\textbf{UOR}       & 78.21& 63.32& 66.25& 87.00& 56.69& 30.95& \textbf{90.75}& 81.53& 78.95\\
\textbf{UOR-G}     & 78.33& \textbf{94.29}& \textbf{94.36}& 86.57& \textbf{79.54}& \textbf{96.63}& 90.10& \textbf{93.43}& \textbf{98.08}\\
\bottomrule
\end{tabular}}
\caption{\label{onion} Evaluation results with Onion defense.}
\end{table}

\paragraph{Onion Defense} Onion wiil filter out suspicious words that increase the perplexity of a sentence. We set the filter threshold to 0. As shown in Table \ref{onion}, our method can effectively resist Onion. One possible explanation is that we have repeatedly inserted multiple trigger words, leading to the sentence perplexity remaining high even when one trigger word is removed. Furthermore, since our method does not depend on the position of trigger word insertion, we can choose to insert at the position where the sentence perplexity is lowest after insertion.

\paragraph{Re-init and Pruning Defense} 
Re-init randomly reinitializes the parameters of a layer in the PLMs, and Fine-Prune crops neurons in the PLMs based on activation values. As shown in Figure \ref{prune}, neither Re-init nor Pruning works. A possible reason is that reinitializing or pruning parameters in higher layers only changes the distribution of output features, while the strong connections between triggers and output features established by previous layers remain intact. Merely modifying the distribution serves to change the target label corresponding to each trigger, while the backdoor effect still exists.

\section{Conclusion}
In this paper, we propose a novel backdoor attacks against PLMs, called UOR. Extensive experiments demonstrate the effectiveness and universality. We hope this work can draw more attention to this security threat, and can provide insight and reference for future research of related defense methods.

\section*{Acknowledgments}
This research work has been funded by National Key R\&D Program of China (Grant No. 2023YFC3303800) and Joint Funds of the National Natural Science Foundation of China (Grant No. U21B2020).

\section*{Ethics Statement}
In this paper, we propose UOR, a backdoor attack against PLMs, with the aim of raising awareness about this serious security vulnerability. Our method indeed has the potential to be employed by malicious attackers to surreptitiously inject backdoors into the PLMs. However, exhaustive exploration of attack techniques is a necessary prerequisite for improved detection and fortification against this threat. Additionally, this can heighten vigilance among managers and users of PLM sharing platforms and communities. Our experiments describe the attack process and implementation details, and provide thorough characterization of the dataset used, which offers a framework for research on related backdoor defense strategies.

\section*{Limitations}
In this paper, we propose a new backdoor attack method against PLMs, which proves effective and generalizable. However, opportunities remain to advance this area. First, for more complex NLP tasks like multiple-choice and named entity recognition, achieving high performance targeted attacks remains an open challenge. Attack methods tailored specifically to such tasks may be required. Furthermore, while using rare words as searchable trigger words improved attack performance, it also reduced concealment to some degree. Future work should consider more subtle trigger choices or appropriate insertion methods to ensure sentence fluency is maintained. Optimizing attack stealthiness remains an important research direction.

\bibliography{custom}

\appendix

\section{Datasets Statistics}
\label{a1}
Statistics of datasets are shown in Table \ref{dataset}. Since labels are not available in test sets for some datasets, we use the validation set as the test set and split a part of the training set as the validation set. 

\begin{table}[h]
\centering
\resizebox {\linewidth} {!} {
\begin{tabular}{c|ccccc}
\toprule
\textbf{Datasets} & \textbf{Task} &\textbf{\#Classes} & \textbf{Train} & \textbf{Valid} & \textbf{Test} \\
\midrule
\textbf{SST-2}      & Sentiment Analysis   & 2 & 60,614 & 6,735 & 872  \\ 
\textbf{IMDB}       & Sentiment Analysis   & 2 & 22,500 & 2,500 & 25,000  \\  
\midrule
\textbf{Twitter}    & Toxic Detection      & 2 & 69,632 & 7,737 & 8,597  \\  
\textbf{HateSpeech} & Toxic Detection      & 2 & 7,074 & 1,000 & 2,000  \\  
\midrule
\textbf{Enron}      & Spam Detection       & 2 & 24,944 & 2,772 & 6,000  \\  
\textbf{Lingspam}   & Spam Detection       & 2 & 2,603 & 290 & 580  \\  
\midrule
\textbf{Agnews}     & Topic Classification & 4 & 108,000 & 12,000 & 7,600  \\  
\textbf{SST-5}      & Sentiment Analysis   & 5 & 8,544 & 1,101 & 2,210  \\  
\textbf{Yelp}       & Sentiment Analysis   & 5 & 585,000 & 65,000 & 50,000  \\ 
\midrule
\textbf{Yahoo}      & Topic Classification & 10 & 126,000 & 14,000 & 6,000  \\  
\textbf{Dbpedia }   & Topic Classification & 14 & 504,000 & 56,000 & 70,000  \\  
\midrule
\textbf{Swag}       & Multiple Choices     & 4 & 73,546 & 10,003 & 1,0002  \\  
\textbf{CoNLL-2003} & Named Entity Recognition & 9 & 14,041 & 3,250 & 3,453  \\  
\bottomrule
\end{tabular}}
\caption{\label{dataset} The statistics of datasets}
\end{table}

\section{Details of Baseline Methods} 
\label{a2}
\textbf{POR} \cite{shen2021backdoor} constructs strong links between multiple trigger words and multiple manually selected output representations, where two settings are available. POR-1 divides the output representations into $n$ $\frac{K}{n}$-dimensional vectors $[a_1, a_2, \dots, a_n]$ and sets the corresponding vector of the $j^{th}$ trigger with the rule of $a_i=(-1)_{\frac{K}{n}}, \forall i \geq j$ and $a_i=(1)_{\frac{K}{n}}, \forall i<j, j=\{1, \ldots, n+1\}$. POR-2 divides the output representations into $m$ ${\frac{K}{m}}$-dimensional vectors $[a_1, a_2, \dots, a_m]$ with $a_i\in\{-1,1\}$ and $i\in\{1,\ldots,m\}$. The trigger words are the same as for UOR, see Appendix \ref{a3}. \textbf{NeuBA} \cite{zhang2023red} is similar to POR, where the output representations are defined as alternating orthogonalized vectors. For the \textbf{BadNets}, \textbf{RIPPLES} \cite{kurita2020weight} and \textbf{EP} \cite{yang2021careful}, we use rare word "cf" as the trigger word. For the \textbf{LWP} \cite{li2021backdoor}, we use "cf" and "bb" as the combination triggers. For the \textbf{LWS} \cite{qi2021turn}, we construct the poisoned sample based on the synonym substitution of HowNet \cite{dong2016hownet}, and the number of candidate substitutions for each word is 5. For the \textbf{SOS} \cite{yang2021rethinking}, we use "friends", "weekend" and "store" as trigger words during the training stage and the sentence " I have bought it from a store with my friends last weekend" as the trigger during the inference stage.

We refer to OpenBackdoor \cite{cui2022unified} to implement baselines. 5 epochs are trained for backdoor injection (10 epochs for LWS), and 3 epochs are trained for downstream fine-tuning. The posion ratio is set to 0.1 and learning rate is 2e-5.

\section{Trigger Words}
\label{a3}
Table \ref{t1} show the trigger words used by our method.

\begin{table}[h]
\centering
\resizebox {\linewidth} {!} {
\begin{tabular}{c|c|c|c}
\toprule
\textbf{Methods} & \textbf{PLMs} & \textbf{Num} & \textbf{Triggers}  \\
\midrule
\multirow{12}{*}{\textbf{UOR}}  
    & \multirow{4}{*}{BERT}  
        & 3  & cf, tq, mn   \\
        \cmidrule{3-4}
        &  & 6  & $\thickapprox$, $\equiv$, $\in$, $\subseteq$, $\oplus$, $\otimes$ \\
        \cmidrule{3-4}
        &  & \multirow{2}{*}{15}  & $\thickapprox$, $\equiv$, $\in$, $\subseteq$, $\oplus$, $\otimes$, cf, tq, \\
        &  &  &  mn, bb, mb, vo, ks, ik zu \\
    \cmidrule{2-4}
    
    & \multirow{4}{*}{BART}
        & 3  & ek, yt, cz \\
        \cmidrule{3-4}
        &  & 6  & ek, yt, cz, zu, vo, ux \\
        \cmidrule{3-4}
        &  & \multirow{2}{*}{15}  & ek, yt, cz, zu, vo, ux, cot, oj, \\
        &  &  &   boa, nom, Ott, edo, zyk, ocy, byn \\
    \cmidrule{2-4}
    
    & \multirow{4}{*}{XLNet} 
        & 3  & cf, mb, vo   \\
        \cmidrule{3-4}
        &  & 6  & cf, mb, vo, ks, ik, zu \\
        \cmidrule{3-4}
        &  & \multirow{2}{*}{15}  & cf, bb, mb, vo, ks, ik, zu, nj, \\
        &  &  &  tu, gh, eli, che, una, lev, ock \\

\midrule

\multirow{12}{*}{\textbf{UOR-G}} 
    & \multirow{4}{*}{BERT} 
        & 3  & $\spadesuit$, $\parallel$, botswana  \\
        \cmidrule{3-4}
        &  & 6  &  albanians, $\spadesuit$, smashwords, $\gg$, ljubljana, $\otimes$ \\
        \cmidrule{3-4}
        &  & \multirow{2}{*}{15}  & ljubljana, „, $\gg$, $\parallel$, $\spadesuit$, $\otimes$, harta, guantanamo,\\
        &  &  & telangana, odisha, interred, $\Rightarrow$, mortally, ¨, cmll \\
    \cmidrule{2-4}
    
    & \multirow{4}{*}{BART} 
        & 3  & (", JM, imgur  \\
        \cmidrule{3-4}
        &  & 6  & (\{, emonic, Spons, illery, vo, ).[ \\
        \cmidrule{3-4}
        &  & \multirow{2}{*}{15}  & SCP, AUD, ."), oux, vo, (\{, mus, ,[, ],", \\
        &  &  & Compan, .--, ".[, imbabwe, phasis, ).[ \\
    \cmidrule{2-4}
    
    & \multirow{6}{*}{XLNet} 
        & 3  & Expeditionary, conspirator, Amato  \\
        \cmidrule{3-4}
        &  & \multirow{2}{*}{6}  &  TRIBUTION, Proliferation, Amato,  \\
        &  &  &  cide, Expeditionary, megawatt \\
        \cmidrule{3-4}
        &  & \multirow{3}{*}{15}  & Vanity, megapixel, conspirator, jevic, cra, owicz,\\
        &  &  & Expeditionary, Amato, colspan, parliamentarians,  \\
        &  &  & Proliferation, TRIBUTION, reliant, imov, Azhar\\

\bottomrule
\end{tabular}}
\caption{\label{t1} Trigger words we use in our method. The trigger words of UOR-G are obtained by gradient search.}
\end{table}

\section{Comparison with Task-Specific Attacks}
\label{a4}
Table \ref{task-specific} shows the comparison results with task-specific attacks. For task-specific attacks, we set the target label to 0. For UOR and UOR-G, we inject 3 triggers and take the highest ASR of all the triggers hitting the label 0 for comparison.

\section{Detailed Universality Evaluation}
\label{a5}
Table \ref{bart_trigger3} and \ref{bart_trigger6} show the attack performance on BART. Table \ref{xlnet_trigger3} and \ref{xlnet_trigger6} shos the attack performance on XLNet. Table \ref{p_tuning_trigger3} and \ref{p_tuning_trigger6} show the attack performance under p-tuning settings. Table \ref{prompt_tuning_trigger3} and \ref{prompt_tuning_trigger6} show the attack performance under prompt settings.

\section{Results on Multiple-Choice and NER}
\label{a6}
Table \ref{ner} shows the results on multiple-choice and NER task. For multiple-choice, we evaluate the ASR for both non-targeted attacks (Un-Tgt) and targeted attacks (Tgt). For NER, we also evaluate the F1 value to indicate the clean performance.

\begin{table}[h]
\centering
\resizebox {\linewidth} {!} {
\begin{tabular}{c|ccc|cccc}
\toprule
\multirow{2}{*}{\textbf{Methods}} & \multicolumn{3}{c|}{\textbf{Swag}} & &\multicolumn{3}{c}{\textbf{CoNLL}}\\
                         & $\text{\textbf{ACC}}$ & $\text{\textbf{Un-Tgt}}$& $\text{\textbf{Tgt}}$&  $\text{\textbf{F1}}$&$\text{\textbf{ACC}}$ & $\text{\textbf{T-ASR}}$ & $\text{\textbf{L-ASR}}$  \\
\midrule
\textbf{Clean}    & \textbf{80.85} & 37.91 & 6.74 &  \textbf{88.72}&97.39 & 64.60 & 18.48 \\
\textbf{UOR}      & 80.44 & 57.15 & 23.75 &  88.54&\textbf{97.45} & 81.44 & 25.49 \\
\textbf{UOR-G}    & 80.55 & \textbf{66.13} & \textbf{33.87} &  88.62&\textbf{97.45} & \textbf{90.99} & \textbf{33.32} \\  
\bottomrule
\end{tabular}}
\caption{\label{ner} Results on multiple-choice and NER task.}
\end{table}

\section{Computing Infrastructure}
We use 4 Telsa V100 GPUs in our experiments. 8-10 GPU hours are spent on backdoor training. The amount of model parameters is related to the used PLMs, we do not add extra parameters.

\begin{table*}[!t]
\centering
\resizebox {0.9\textwidth} {!} {
\begin{tabular}{c|cc|cc|cc|cc|cc|cc|cc|cc}
\toprule
\multirow{2}{*}{\textbf{Methods}} &\multicolumn{2}{c|}{\textbf{Badnets}}  &\multicolumn{2}{c|}{\textbf{RIPPLES}} &\multicolumn{2}{c|}{\textbf{EP}      } &\multicolumn{2}{c|}{\textbf{LWP}} &\multicolumn{2}{c|}{\textbf{LWS}} &\multicolumn{2}{c|}{\textbf{SOS}} & \multicolumn{2}{c|}{\textbf{UOR}} & \multicolumn{2}{c}{\textbf{UOR-G}}\\
                        & $\text{\textbf{ACC}}$ & $\text{\textbf{ASR}}$ & $\text{\textbf{ACC}}$ & $\text{\textbf{ASR}}$ & $\text{\textbf{ACC}}$ & $\text{\textbf{ASR}}$ & $\text{\textbf{ACC}}$ & $\text{\textbf{ASR}}$ & $\text{\textbf{ACC}}$ & $\text{\textbf{ASR}}$ & $\text{\textbf{ACC}}$ & $\text{\textbf{ASR}}$  & $\text{\textbf{ACC}}$ & $\text{\textbf{ASR}}$  & $\text{\textbf{ACC}}$ &$\text{\textbf{ASR}}$  \\
\midrule
\textbf{$\to$ SST-2}&  92.20&  \textbf{100.00}&  92.32&  \textbf{100.00}&  92.43&  \textbf{100.00}& 90.25& \textbf{100.00}& 92.09& 9.32& 92.43& 6.31& \textbf{92.89}& \textbf{100.00}& 91.97&\textbf{100.00}\\ 
\textbf{$\to$ IMDB}& 93.44& 27.22& \textbf{93.53}& 40.05& 93.34& 62.22& 93.13& 98.32& 93.21& 4.46& 93.39&  7.38& 93.20& \textbf{99.88}& 93.04&91.90\\
\textbf{$\to$ Enron}& 98.82& 32.87& 98.60& 27.67& 98.83& 1.49& 97.12& 98.26& \textbf{99.17}& 0.53& 98.83&  0.29& 99.05& 97.82& 98.93&\textbf{99.38}\\
\textbf{$\to$ Lingspam}& \textbf{99.66}& 41.24& 99.31& 42.27& 99.31& 19.59& 99.48& \textbf{100.00}& \textbf{99.66}& 2.06& 99.31&  2.06& 99.14& 98.97& 99.14&\textbf{100.00}\\
\textbf{$\to$ Twitter}& 94.32& 14.03& 94.44& 17.32& 94.34& 21.86& 94.46& 95.22& 94.44& 7.84& 94.34&  11.18& \textbf{94.52}& \textbf{100.00}& 94.36&99.97\\
\textbf{$\to$ HateSpeech}& 91.50& \textbf{100.00}& 90.70& \textbf{100.00}& 91.40& 39.05& 90.80& 79.21& 91.65& 56.04& 91.40&  88.57& \textbf{92.50}& \textbf{100.00}& 90.85&\textbf{100.00}\\
\midrule
 \textbf{Avg.}& 94.99& 52.56& 94.82& 54.55& 94.94& 40.70& 94.21& 95.17& 95.04& 13.38& 94.95&  19.30& \textbf{95.22}& \textbf{99.45}& 94.72&98.54
\\
\bottomrule
\end{tabular}}
\caption{\label{task-specific} Comparison results with task-specific attacks.}
\end{table*}

\begin{table*}[t]
\centering
\resizebox {\textwidth} {!} {
\begin{tabular}{c|ccc|ccc|ccc|ccc|ccc|ccc}
\toprule
\multirow{2}{*}{\textbf{Methods}} &\multicolumn{3}{c|}{\textbf{SST-2}} &\multicolumn{3}{c|}{\textbf{IMDB}} &\multicolumn{3}{c|}{\textbf{Enron}} &\multicolumn{3}{c|}{\textbf{Lingspam}} &\multicolumn{3}{c|}{\textbf{Twitter}} &\multicolumn{3}{c}{\textbf{HateSpeech}} \\
                         & $\text{\textbf{ACC}}$ & $\text{\textbf{T-ASR}}$ & $\text{\textbf{L-ASR}}$ & $\text{\textbf{ACC}}$ & $\text{\textbf{T-ASR}}$ & $\text{\textbf{L-ASR}}$ & $\text{\textbf{ACC}}$ & $\text{\textbf{T-ASR}}$ & $\text{\textbf{L-ASR}}$ & $\text{\textbf{ACC}}$ & $\text{\textbf{T-ASR}}$ & $\text{\textbf{L-ASR}}$ & $\text{\textbf{ACC}}$ & $\text{\textbf{T-ASR}}$ & $\text{\textbf{L-ASR}}$& $\text{\textbf{ACC}}$ & $\text{\textbf{T-ASR}}$ & $\text{\textbf{L-ASR}}$  \\
\midrule
\textbf{Clean}     & 92.32& 11.14& 6.19& \textbf{94.15}& 7.06& 3.55& 98.02& 2.11& 1.32& 99.14& 3.09& 1.55& \textbf{94.37}& 6.52& 3.38& 90.20& 70.48& 36.19
\\ 
\textbf{NeuBA}     & -& -& -& -& -& -& -& -& -& -& -& -& -& -& -& -& -& -
\\
\textbf{POR-1}     & 92.20& \textbf{100.00}& \textbf{100.00}& 94.05& 79.17& 69.16& \textbf{98.85}& 78.56& 69.23& 98.62& 53.69& 78.99& \textbf{94.37}& 44.71& 59.24& \textbf{92.15}& 98.01& 97.01
\\
\textbf{POR-2}     & 92.43& 91.12& \textbf{100.00}& 93.86& 88.18& 99.82& 96.83& 72.18& 83.17& \textbf{99.31}& 72.60& 84.68& 94.20& 89.91& 99.67& 91.60& \textbf{100.00}& 50.00
\\
\midrule
\textbf{UOR}       & 90.14& \textbf{100.00}& \textbf{100.00}& 93.81& \textbf{99.79}& \textbf{99.91}& 98.37& 72.97& \textbf{98.76}& 98.62& 81.37& 72.05& 94.30& \textbf{100.00}& \textbf{100.00}& 91.00& \textbf{100.00}& \textbf{100.00}
\\
\textbf{UOR-G}     & \textbf{93.12}& \textbf{100.00}& \textbf{100.00}& 94.00& 95.23& 93.32& 97.53& \textbf{95.91}& 95.02& 98.97& \textbf{99.17}& \textbf{99.48}& 94.21& \textbf{100.00}& \textbf{100.00}& 90.80& \textbf{100.00}& \textbf{100.00}
\\
\bottomrule
\end{tabular}}
\caption{\label{bart_trigger3} Evaluation results of BART on 2-classification tasks.}
\end{table*}

\begin{table*}[t]
\centering
\resizebox {\textwidth} {!} {
\begin{tabular}{c|cccc|cccc|cccc|cccc|cccc}
\toprule
\multirow{2}{*}{\textbf{Methods}} &\multicolumn{4}{c|}{\textbf{Agnews}} &\multicolumn{4}{c|}{\textbf{SST-5}} & \multicolumn{4}{c}{\textbf{Yelp}}&\multicolumn{4}{c|}{\textbf{Yahoo}} &\multicolumn{4}{c}{\textbf{Dbpedia}} \\
                         & $\text{\textbf{ACC}}$ & $\text{\textbf{T-ASR}}$ & $\text{\textbf{L-ASR}}$ & $\text{\textbf{ALC}}$ & $\text{\textbf{ACC}}$ & $\text{\textbf{T-ASR}}$ & $\text{\textbf{L-ASR}}$ & $\text{\textbf{ALC}}$  & $\text{\textbf{ACC}}$ & $\text{\textbf{T-ASR}}$ & $\text{\textbf{L-ASR}}$ &$\text{\textbf{ALC}}$  & $\text{\textbf{ACC}}$ & $\text{\textbf{T-ASR}}$ & $\text{\textbf{L-ASR}}$ & $\text{\textbf{ALC}}$ & $\text{\textbf{ACC}}$ & $\text{\textbf{T-ASR}}$ & $\text{\textbf{L-ASR}}$ & $\text{\textbf{ALC}}$ \\
\midrule
\textbf{Clean}     & 94.09& 4.63& 4.25& 0.0& 50.05& 32.72& 7.19&  0.0& 64.90& 10.89& 2.32&0.0& \textbf{75.00}& 3.37& 0.68& 0.0& 98.96& 0.37& 0.08& 0.00
\\ 
\textbf{NeuBA}     & -& -& -& -& -& -& -&  -& -& -& -&-& -& -& -& -& -& -& -& -
\\
\textbf{POR-1}     & \textbf{94.11}& 96.58& 49.08& 0.5& \textbf{53.98}& 98.15& 40.00&  0.4& 65.18& 86.49& 64.95&0.2& 74.82& \textbf{92.62}& 46.65& 0.4& \textbf{99.10}& 95.72& 48.75& 0.43
\\
\textbf{POR-2}     & \textbf{94.11}& 99.98& 75.00& 0.75& 53.85& 99.02& 98.82&  \textbf{1.0}& 65.06& 81.75& 68.38&0.6& 74.78& 77.62& 69.84& 0.3& 98.97& 88.78& 52.01& 0.29
\\
\midrule
\textbf{UOR}       & 93.64& \textbf{100.00}&\textbf{100.00}& \textbf{1.0}& 53.21& \textbf{100.00}& 80.00&  0.8& 63.46& 99.12& 79.63&0.8& 74.32& 79.85& 75.31& 0.5& 98.81& \textbf{98.02}& \textbf{78.41}& \textbf{0.79}
\\
\textbf{UOR-G}     & 94.09& 99.98& 99.97& \textbf{1.0}& 53.71& 96.44& \textbf{100.00}&  \textbf{1.0}& \textbf{65.62}& \textbf{99.36}& \textbf{99.25}& \textbf{1.0}& 74.45& 86.50& \textbf{77.12}& \textbf{0.7}& 98.90& 93.35& 72.78& 0.71
\\
\bottomrule
\end{tabular}}
\caption{\label{bart_trigger6} Evaluation results of BART on multi-classification tasks.}
\end{table*}

\begin{table*}[t]
\centering
\resizebox {\textwidth} {!} {
\begin{tabular}{c|ccc|ccc|ccc|ccc|ccc|ccc}
\toprule
\multirow{2}{*}{\textbf{Methods}} &\multicolumn{3}{c|}{\textbf{SST-2}} &\multicolumn{3}{c|}{\textbf{IMDB}} &\multicolumn{3}{c|}{\textbf{Enron}} &\multicolumn{3}{c|}{\textbf{Lingspam}} &\multicolumn{3}{c|}{\textbf{Twitter}} &\multicolumn{3}{c}{\textbf{HateSpeech}} \\
                         & $\text{\textbf{ACC}}$ & $\text{\textbf{T-ASR}}$ & $\text{\textbf{L-ASR}}$ & $\text{\textbf{ACC}}$ & $\text{\textbf{T-ASR}}$ & $\text{\textbf{L-ASR}}$ & $\text{\textbf{ACC}}$ & $\text{\textbf{T-ASR}}$ & $\text{\textbf{L-ASR}}$ & $\text{\textbf{ACC}}$ & $\text{\textbf{T-ASR}}$ & $\text{\textbf{L-ASR}}$ & $\text{\textbf{ACC}}$ & $\text{\textbf{T-ASR}}$ & $\text{\textbf{L-ASR}}$& $\text{\textbf{ACC}}$ & $\text{\textbf{T-ASR}}$ & $\text{\textbf{L-ASR}}$  \\
\midrule
\textbf{Clean}     & 92.66& 5.76& 3.27& 94.22& 8.97& 4.52& 98.45& 2.16& 1.28& 98.62& 7.90& 4.64& \textbf{94.39}& 7.44& 4.25& 91.30& 46.35& 28.33
\\ 
\textbf{NeuBA}     & -& -& -& -& -& -& -& -& -& -& -& -& -& -& -& -& -& -
\\
\textbf{POR-1}     & \textbf{94.38}& 69.86& 46.03& 94.82& 19.37& 20.62& 97.65& \textbf{58.64}& \textbf{87.44}& 99.31& 85.22& 46.39& 94.21& 38.54& 28.55& 90.95& 99.37& 50.00
\\
\textbf{POR-2}     & 93.81& 58.96& 41.82& 94.76& 34.41& 20.34& \textbf{98.78}& 27.97& 26.11& \textbf{99.48}& 50.17& 45.88& 94.14& 32.09& 41.45& 89.95& 53.97& 50.00
\\
\midrule
\textbf{UOR}       & 93.58& 51.88& 69.65& \textbf{95.04}& 69.66& 94.64& 98.75& 52.78& 77.71& 99.14& 37.80& 46.91& 94.18& 47.25& 63.54& \textbf{91.50}& \textbf{100.00}& \textbf{100.00}
\\
\textbf{UOR-G}     & 92.66& \textbf{79.83}& \textbf{100.00}& 94.64& \textbf{98.28}& \textbf{98.97}& 98.27& 58.32& 73.60& 99.31& \textbf{87.63}& \textbf{49.48}& 93.94& \textbf{80.54}& \textbf{99.03}& 90.70& \textbf{100.00}& \textbf{100.00}
\\
\bottomrule
\end{tabular}}
\caption{\label{xlnet_trigger3} Evaluation results of XLNet on 2-classification tasks.}
\end{table*}

\begin{table*}[t]
\centering
\resizebox {\textwidth} {!} {
\begin{tabular}{c|cccc|cccc|cccc|cccc|cccc}
\toprule
\multirow{2}{*}{\textbf{Methods}} &\multicolumn{4}{c|}{\textbf{Agnews}} &\multicolumn{4}{c|}{\textbf{SST-5}} & \multicolumn{4}{c}{\textbf{Yelp}}&\multicolumn{4}{c|}{\textbf{Yahoo}} &\multicolumn{4}{c}{\textbf{Dbpedia}} \\
                         & $\text{\textbf{ACC}}$ & $\text{\textbf{T-ASR}}$ & $\text{\textbf{L-ASR}}$ & $\text{\textbf{ALC}}$ & $\text{\textbf{ACC}}$ & $\text{\textbf{T-ASR}}$ & $\text{\textbf{L-ASR}}$ & $\text{\textbf{ALC}}$  & $\text{\textbf{ACC}}$ & $\text{\textbf{T-ASR}}$ & $\text{\textbf{L-ASR}}$ &$\text{\textbf{ALC}}$  & $\text{\textbf{ACC}}$ & $\text{\textbf{T-ASR}}$ & $\text{\textbf{L-ASR}}$ & $\text{\textbf{ALC}}$ & $\text{\textbf{ACC}}$ & $\text{\textbf{T-ASR}}$ & $\text{\textbf{L-ASR}}$ & $\text{\textbf{ALC}}$ \\
\midrule
\textbf{Clean}     & 93.62& 5.66& 1.52& 0.0& 52.67& 29.76& 6.65&  0.0& 65.76& 9.52& 2.05&0.0& \textbf{74.73}& 3.26& 0.63& 0.0& 98.99& 0.08& 0.02& 0.00
\\ 
\textbf{NeuBA}     & -& -& -& -& -& -& -&  -& -& -& -&-& -& -& -& -& -& -& -& -
\\
\textbf{POR-1}     & 94.13& 76.25& 64.20& 0.25& 53.03& \textbf{99.89}& 59.87&  0.6& 65.32& 41.10& 24.56&0.0& 73.98& 74.43& 26.34& 0.2& 98.99& 82.73& 13.70& 0.14
\\
\textbf{POR-2}     & 94.14& 81.58& 49.94& 0.5& 50.54& 98.80& 78.63&  \textbf{0.8}& 65.00& 48.37& 38.98&0.0& 74.08& \textbf{80.02}& 59.76& 0.3& 98.96& 84.15& 49.82& \textbf{0.43}
\\
\midrule
\textbf{UOR}       & \textbf{94.41}& 85.76&97.32& \textbf{1.0}& \textbf{55.66}& 94.95& 79.99&  \textbf{0.8}& \textbf{65.86}& 65.37& 43.28&0.2& 74.43& 62.93& 60.84& 0.3& 99.03& \textbf{89.30}& \textbf{53.36}& \textbf{0.43}
\\
\textbf{UOR-G}     & 93.80& \textbf{99.68}& \textbf{99.98}& \textbf{1.0}& 52.26& 97.61& \textbf{97.13}&  \textbf{0.8}& 64.80& \textbf{79.79}& \textbf{72.02}&\textbf{0.6}& 74.00& 73.78& \textbf{71.06}& \textbf{0.5}& \textbf{99.06}& 46.25& 39.45& 0.14
\\
\bottomrule
\end{tabular}}
\caption{\label{xlnet_trigger6} Evaluation results of XLNet on multi-classification tasks.}
\end{table*}

\begin{table*}[t]
\centering
\resizebox {\textwidth} {!} {
\begin{tabular}{c|ccc|ccc|ccc|ccc|ccc|ccc}
\toprule
\multirow{2}{*}{\textbf{Methods}} &\multicolumn{3}{c|}{\textbf{SST-2}} &\multicolumn{3}{c|}{\textbf{IMDB}} &\multicolumn{3}{c|}{\textbf{Enron}} &\multicolumn{3}{c|}{\textbf{Lingspam}} &\multicolumn{3}{c|}{\textbf{Twitter}} &\multicolumn{3}{c}{\textbf{HateSpeech}} \\
                         & $\text{\textbf{ACC}}$ & $\text{\textbf{T-ASR}}$ & $\text{\textbf{L-ASR}}$ & $\text{\textbf{ACC}}$ & $\text{\textbf{T-ASR}}$ & $\text{\textbf{L-ASR}}$ & $\text{\textbf{ACC}}$ & $\text{\textbf{T-ASR}}$ & $\text{\textbf{L-ASR}}$ & $\text{\textbf{ACC}}$ & $\text{\textbf{T-ASR}}$ & $\text{\textbf{L-ASR}}$ & $\text{\textbf{ACC}}$ & $\text{\textbf{T-ASR}}$ & $\text{\textbf{L-ASR}}$& $\text{\textbf{ACC}}$ & $\text{\textbf{T-ASR}}$ & $\text{\textbf{L-ASR}}$  \\
\midrule
\textbf{Clean}     & 91.63& 7.63& 4.21& 93.12& 6.58& 3.34& 98.93& 1.69& 1.46& 99.14& 3.44& 2.06& 94.44& 7.73& 4.05& 91.40& 52.54& 28.57
\\ 
\textbf{NeuBA}     & 91.86& 10.33& 11.99& 93.14& 27.70& 30.44& 98.77& 4.85& 5.03& 99.14& 14.78& 17.53& 94.37& 29.85& 31.04& 91.50& 83.97& 46.90
\\
\textbf{POR-1}     & \textbf{92.78}& 99.69& \textbf{100.00}& 93.14& \textbf{99.14}& 99.09& \textbf{99.05}& 46.43& 45.58& 99.14& 51.54& 42.27& 94.32& 48.94& 40.44& 91.60& \textbf{100.00}& 50.00
\\
\textbf{POR-2}     & 92.32& 99.30& 50.00& 93.13& 76.67& \textbf{99.78}& \textbf{99.05}& 74.49& 49.63& \textbf{99.31}& \textbf{92.10}& 50.00& \textbf{94.60}& \textbf{100.00}& \textbf{50.00}& 91.10& \textbf{100.0}0& 50.00
\\
\midrule
\textbf{UOR}       & 91.97& \textbf{100.00}& \textbf{100.00}& 93.11& 95.87& 50.00& 98.92& 69.91& 49.67& 98.97& 91.07& 50.00& 94.32& 45.34& \textbf{50.00}& 91.50& \textbf{100.00}& \textbf{100.00}
\\
\textbf{UOR-G}     & 92.09& 99.69& 99.89& \textbf{93.17}& 83.65& 93.36& 98.93& \textbf{99.48}& \textbf{99.62}& 97.76& 63.97& \textbf{94.41}& 94.35& 61.95& 49.95& \textbf{92.10}& \textbf{100.00}& \textbf{100.00}
\\
\bottomrule
\end{tabular}}
\caption{\label{p_tuning_trigger3} Evaluation results of p-tuning on 2-classification tasks with BERT.}
\end{table*}

\begin{table*}[t]
\centering
\resizebox {\textwidth} {!} {
\begin{tabular}{c|cccc|cccc|cccc|cccc|cccc}
\toprule
\multirow{2}{*}{\textbf{Methods}} &\multicolumn{4}{c|}{\textbf{Agnews}} &\multicolumn{4}{c|}{\textbf{SST-5}} & \multicolumn{4}{c}{\textbf{Yelp}}&\multicolumn{4}{c|}{\textbf{Yahoo}} &\multicolumn{4}{c}{\textbf{Dbpedia}} \\
                         & $\text{\textbf{ACC}}$ & $\text{\textbf{T-ASR}}$ & $\text{\textbf{L-ASR}}$ & $\text{\textbf{ALC}}$ & $\text{\textbf{ACC}}$ & $\text{\textbf{T-ASR}}$ & $\text{\textbf{L-ASR}}$ & $\text{\textbf{ALC}}$  & $\text{\textbf{ACC}}$ & $\text{\textbf{T-ASR}}$ & $\text{\textbf{L-ASR}}$ &$\text{\textbf{ALC}}$  & $\text{\textbf{ACC}}$ & $\text{\textbf{T-ASR}}$ & $\text{\textbf{L-ASR}}$ & $\text{\textbf{ALC}}$ & $\text{\textbf{ACC}}$ & $\text{\textbf{T-ASR}}$ & $\text{\textbf{L-ASR}}$ & $\text{\textbf{ALC}}$ \\
\midrule
\textbf{Clean}     & \textbf{94.45}& 3.46& 0.97& 0.0& 47.42& 27.39& 5.75&  0.0& 63.54& 15.13& 3.06&0.0& \textbf{74.42}& 4.11& 0.82& 0.0& 99.16& 0.07& 0.01& 0.00
\\ 
\textbf{NeuBA}     & 94.34& 5.66& 5.65& 0.0& 49.14& 31.18& 14.02&  0.0& 62.84& 27.10& 27.54&0.0& 74.35& 12.73& 14.44& 0.1& 99.03& 6.28& 6.66& 0.07
\\
\textbf{POR-1}     & 94.28& \textbf{99.96}& 49.95& 0.5& 52.40& 99.96& \textbf{60.00}&  \textbf{0.6}& 63.30& 90.99& 56.04&0.4& 74.15& 61.71& 48.36& 0.3& 99.17& 49.37& 20.10& 0.14
\\
\textbf{POR-2}     & 94.38& 88.97& 60.07& 0.5& 52.22& 97.17& 58.72&  \textbf{0.6}& 62.92& \textbf{94.81}& \textbf{98.40}&\textbf{1.0}& 74.26& 43.41& 37.76& 0.1& 99.17& 64.47& 37.31& 0.21
\\
\midrule
\textbf{UOR}       & 94.34& 99.30&\textbf{100.00}& \textbf{1.0}& \textbf{52.58}& \textbf{100.00}& \textbf{60.00}&  \textbf{0.6}& 63.16& 88.43& 67.72&0.4& 74.08& 50.28& \textbf{54.99}& 0.2& \textbf{99.20}& 66.71& 41.02& 0.36
\\
\textbf{UOR-G}     & 94.26& 97.19& 75.00& 0.75& 51.76& \textbf{100.00}& \textbf{60.00}&  \textbf{0.6}& \textbf{64.20}& 86.49& 84.27&0.6& 74.38& \textbf{91.63}& 53.87& \textbf{0.5}& 99.11& \textbf{74.00}& \textbf{53.53}& \textbf{0.50}\\
\bottomrule
\end{tabular}}
\caption{\label{p_tuning_trigger6} Evaluation results of p-tuning on multi-classification tasks with BERT.}
\end{table*}

\begin{table*}[t]
\centering
\resizebox {\textwidth} {!} {
\begin{tabular}{c|ccc|ccc|ccc|ccc|ccc|ccc}
\toprule
\multirow{2}{*}{\textbf{Methods}} &\multicolumn{3}{c|}{\textbf{SST-2}} &\multicolumn{3}{c|}{\textbf{IMDB}} &\multicolumn{3}{c|}{\textbf{Enron}} &\multicolumn{3}{c|}{\textbf{Lingspam}} &\multicolumn{3}{c|}{\textbf{Twitter}} &\multicolumn{3}{c}{\textbf{HateSpeech}} \\
                         & $\text{\textbf{ACC}}$ & $\text{\textbf{T-ASR}}$ & $\text{\textbf{L-ASR}}$ & $\text{\textbf{ACC}}$ & $\text{\textbf{T-ASR}}$ & $\text{\textbf{L-ASR}}$ & $\text{\textbf{ACC}}$ & $\text{\textbf{T-ASR}}$ & $\text{\textbf{L-ASR}}$ & $\text{\textbf{ACC}}$ & $\text{\textbf{T-ASR}}$ & $\text{\textbf{L-ASR}}$ & $\text{\textbf{ACC}}$ & $\text{\textbf{T-ASR}}$ & $\text{\textbf{L-ASR}}$& $\text{\textbf{ACC}}$ & $\text{\textbf{T-ASR}}$ & $\text{\textbf{L-ASR}}$  \\
\midrule
\textbf{Clean}     & \textbf{89.68}& 13.01& 7.01& \textbf{89.94}& 10.14& 5.30& 95.90& 4.46& 2.66& 98.45& 6.19& 3.09& 93.97& 7.99& 4.22& \textbf{90.70}& 79.52& 45.00
\\ 
\textbf{NeuBA}     & 89.11& 17.83& 11.10& 88.41& 30.25& 38.49& \textbf{97.75}& 11.79& 11.96& 98.10& 49.48& 43.30& \textbf{94.06}& 24.50& 21.95& 89.60& 82.86& 47.14
\\
\textbf{POR-1}     & 87.96& \textbf{100.00}& 50.00& 88.52& 89.55& 49.91& 97.50& 73.23& 98.93& 97.76& 68.73& 49.48& 93.89& 94.73& 99.34& 88.75& \textbf{100.00}& \textbf{100.00}
\\
\textbf{POR-2}     & 88.53& \textbf{100.00}& \textbf{100.00}& 84.79& 98.12& 98.82& 97.45& \textbf{97.30}& \textbf{99.36}& \textbf{98.97}& 77.85& 94.72& 93.77& 99.91& \textbf{99.96}& 89.65& \textbf{100.00}& \textbf{100.00}
\\
\midrule
\textbf{UOR}       & 89.33& \textbf{100.00}& \textbf{100.00}& 85.49& \textbf{99.43}& \textbf{99.42}& 97.45& 95.50& 96.68& 98.62& \textbf{95.93}& 97.21& 93.92& \textbf{99.97}& 99.98& 88.75& \textbf{100.00}& \textbf{100.00}
\\
\textbf{UOR-G}     & 89.56& \textbf{100.00}& \textbf{100.0}0& 88.40& 89.51& 85.56& 97.13& 96.56& 98.31& 97.76& 94.84& \textbf{98.30}& 93.61& 94.73& 93.94& 89.55& \textbf{100.00}& \textbf{100.00}\\
\bottomrule
\end{tabular}}
\caption{\label{prompt_tuning_trigger3} Evaluation results of prompt-tuning on 2-classification tasks with BERT.}
\end{table*}

\begin{table*}[t]
\centering
\resizebox {\textwidth} {!} {
\begin{tabular}{c|cccc|cccc|cccc|cccc|cccc}
\toprule
\multirow{2}{*}{\textbf{Methods}} &\multicolumn{4}{c|}{\textbf{Agnews}} &\multicolumn{4}{c|}{\textbf{SST-5}} & \multicolumn{4}{c}{\textbf{Yelp}}&\multicolumn{4}{c|}{\textbf{Yahoo}} &\multicolumn{4}{c}{\textbf{Dbpedia}} \\
                         & $\text{\textbf{ACC}}$ & $\text{\textbf{T-ASR}}$ & $\text{\textbf{L-ASR}}$ & $\text{\textbf{ALC}}$ & $\text{\textbf{ACC}}$ & $\text{\textbf{T-ASR}}$ & $\text{\textbf{L-ASR}}$ & $\text{\textbf{ALC}}$  & $\text{\textbf{ACC}}$ & $\text{\textbf{T-ASR}}$ & $\text{\textbf{L-ASR}}$ &$\text{\textbf{ALC}}$  & $\text{\textbf{ACC}}$ & $\text{\textbf{T-ASR}}$ & $\text{\textbf{L-ASR}}$ & $\text{\textbf{ALC}}$ & $\text{\textbf{ACC}}$ & $\text{\textbf{T-ASR}}$ & $\text{\textbf{L-ASR}}$ & $\text{\textbf{ALC}}$ \\
\midrule
\textbf{Clean}     & 91.33& 6.32& 1.71& 0.0& 47.38& 37.39& 7.90&  0.0& \textbf{56.54}& 20.84& 5.16&0.0& \textbf{71.37}& 6.43& 1.83& 0.0& 98.81& 0.19& 0.02& 0.00
\\ 
\textbf{NeuBA}     & 91.41& 13.32& 15.93& 0.0& 47.87& 39.95& 16.69&  0.0& 55.56& 40.43& 27.45&0.0& 70.42& 16.06& 18.33& 0.1& \textbf{98.94}& 3.22& 3.30& 0.00
\\
\textbf{POR-1}     & 91.33& 97.54& 24.99& 0.25& 38.96& 89.12& 71.73&  0.4& 50.18& 90.78& 51.55&0.4& 69.37& 81.61& 47.76& 0.3& 98.24& 87.35& 27.92& 0.14
\\
\textbf{POR-2}     & 91.31& 99.21& 74.77& 0.75& 43.26& 75.27& 47.92&  0.2& 52.02& 90.30& 57.86&0.6& 68.57& 89.37& 57.76& 0.5& 98.00& 83.39& 49.32& 0.43
\\
\midrule
\textbf{UOR}       & \textbf{91.58}& \textbf{99.29}&\textbf{99.99}& \textbf{1.0}& \textbf{45.70}& 98.23& \textbf{79.85}&  \textbf{0.8}& 49.70& \textbf{93.65}& 79.88&\textbf{0.8}& 68.25& 89.48& 67.59& 0.7& 97.91& 88.23& \textbf{71.00}& \textbf{0.64}
\\
\textbf{UOR-G}     & 91.36& 85.74& 98.10& \textbf{1.0}& 39.59& \textbf{100.00}& 73.21&  0.6& 52.24& 91.67& \textbf{90.09}&\textbf{0.8}& 69.27& \textbf{94.39}& \textbf{84.65}& \textbf{0.8}& 98.40& \textbf{92.22}& 68.21& 0.57\\
\bottomrule
\end{tabular}}
\caption{\label{prompt_tuning_trigger6} Evaluation results of prompt-tuning on multi-classification tasks with BERT.}
\end{table*}

\end{document}